# Copy Move Forgery using Hu's Invariant Moments and Log-Polar Transformations

Tejas K , Swathi C, Rajesh Kumar M, *Senior member, IEEE*
School of Electronics Engineering
Vellore Institute of Technology
Vellore, India
tejas.krishnareddy1415@gmail.com, chintala.swathi14@gmail.com,
mrajeshkumar@vit.ac.in

*Abstract*— With the increase in interchange of data, there is a growing necessity of security. Considering the volumes of digital data that is transmitted, they are in need to be secure. Among the many forms of tampering possible, one widespread technique is Copy Move Forgery (CMF). This forgery occurs when parts of the image are copied and duplicated elsewhere in the same image. There exist a number of algorithms to detect such a forgery in which the primary step involved is feature extraction. The feature extraction techniques employed must have lesser time and space complexity involved for an efficient and faster processing of media. Also, majority of the existing state of art techniques often tend to falsely match similar genuine objects as copy move forged during the detection process. To tackle these problems, the paper proposes a novel algorithm that recognizes a unique approach of using Hu's Invariant Moments and Log-polar Transformations to reduce feature vector dimension to one feature per block simultaneously detecting CMF among genuine similar objects in an image. The qualitative and quantitative results obtained demonstrate the effectiveness of this algorithm.

## I. INTRODUCTION

Over the past years, there are several detection techniques evolved to authenticate digital media. These detection techniques can be broadly classified into Active methods and Passive methods. Active methods are the one's which are used to detect latent information in a media such as digital watermarking or signatures. This hidden information can be later used to locate the source of such an image or detect potential forgery in the subjected image. Passive methods are complex and advanced methods that use various methods to extract the binary information in the image to find possible traces of tampering in it.

Among passive image forgery detection techniques CMFD is one of the common topics of research recently. Copy move forgery is a forgery where a group of pixels are copied from a region and are pasted in another part of the same image to shroud some important data. Since, the copied part exists in the same image unlike image splicing, certain principle properties stay intact that can be used to detect this kind of forgery in an image. Broadly, the process of CMFD can be summarized by the steps illustrated in Fig 1. A given image of size M x N is divided into overlapping blocks of size B x B. Important features are extracted from each of these blocks by various feature extraction algorithms. These extracted values of each block are stored as linear rows of a new matrix whose size equates to [(M-B)*(N-B)] x N. Further two columns are added to this matrix delineating the location of the first pixel of the corresponding block features. The rows are sorted lexicographically and rows are subjected to undergo a check of similarity. If adjacent rows are found to be similar, then threshold is applied to the Euclidean distance between matching blocks to reduce the number of false positives. The blocks that are lie in the user-specified threshold region are marked to be copy moved.

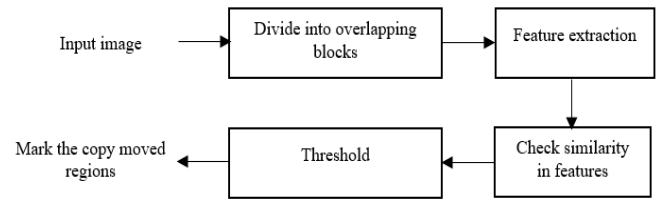

Fig 1: Generalized Block Diagram of Copy Move forgery Detection

In order to reduce the computational complexity in the detection process, several feature extraction techniques have been evolved over the years. But, the scope for reducing the complexity still exists. Also, majority of these algorithms generally tend to confuse between actual copy moved regions and genuine similar regions in an image such as identical windows or two similar products of the same brand during the detection process. In this paper, we aim to propose a novel algorithm that can tackle the above described limitations.

## II. LITERATURE REVIEW

Fridrich et al [1] proposed the first CMFD algorithm using exact match technique where every pixel was counted as a feature and robust CMFD algorithm using DCT coefficients as features of the blocks. Huang et al [2] improved the DCT algorithm to compute the results faster. Farid and Popescu [3] proposed an algorithm to detect CMFD using considerably less feature vector dimension using Principle Component Analysis (PCA) algorithm. Kang et al [4] proposed an algorithm to curb copy move forgery using Singular Value Decomposition (SVD) algorithm which was effectively robust against induced noise. Zhang et al [5] used Discreet Wavelet Transform (DWT) to reduce the complexity of the program as compared to the other existing schemes. Yang et al [6] used Dydadic Wavelet Transform by decomposing the forged image into four sub-bands and removing the low frequency components in it. Muhammad et al [7] proposed a similar algorithm using DyWT which was capable of utilizing both low and high frequency components in an image to eliminate as many false positives as possible. Rahul et al [8] proposed a blur invariant CMFD technique using SWT-SVD algorithm. A method using Fourier Mellin Transform was developed in [9] which proved to be efficient in detecting forgery in highly compressed images. Guangjie et al [10] proposed an algorithm using Hu's invariant moments [11], proving its robustness against several post processing techniques. Huang et al [12] proposed an algorithm using DWT and SVD for robust feature extraction. The PCA algorithm was further developed by Sunil et al [13] to increase its robustness to JPEG compression and noise using DCT-PCA algorithms. PCA is mainly used to reduce the feature vector dimension in the given matrix.

### III. PROPOSED METHOD

We now propose a novel algorithm to reduce the feature vector dimension and simultaneously making the algorithm more effective in differentiating between similar objects in image and actual copy move forgery detection using Hu's invariant moments and log-polar transformations. This algorithm can be discussed in detail as follows:

**Algorithm 1:**

**Input:** Copy Move Forged image.
**Output:** Binary image showing the regions of duplication.
1. Input the forged image of size M x N, convert it to grey scale.
2. Divide the image into overlapping blocks of size B x B.
3. Calculate Hu's invariant moments for each of the divided blocks in step 2 up to $7^{th}$ order.
4. Apply the log-polar transformation over each of the Hu's invariant moment order.
5. Use 'format long' in MATLAB to check on every value up to its respective $15^{th}$ decimal.
6. Calculate the sum of all 7 invariant moments produced for each block and write this value into a new linear column matrix.
7. Add two additional columns to the matrix formed in step 5 indicating the location of the corresponding block's first pixel.

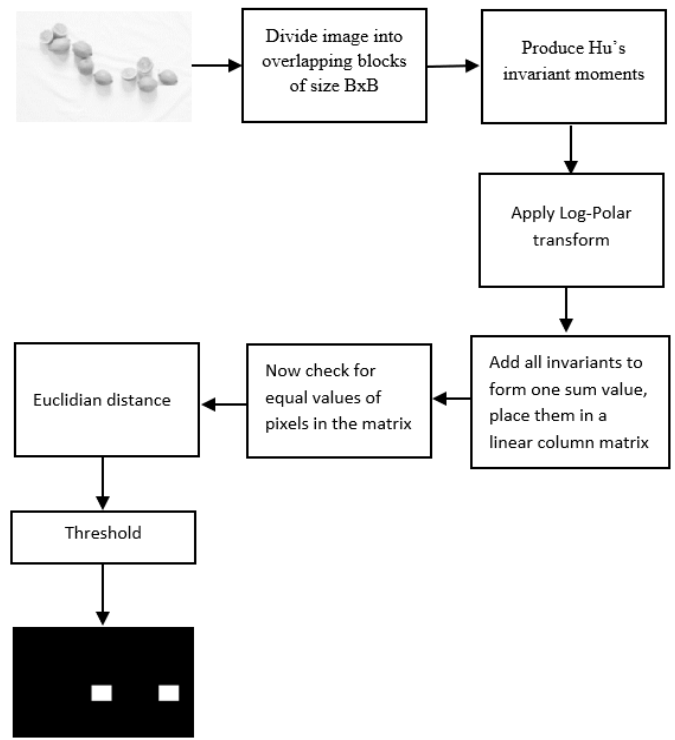

Fig. 2 Proposed Block Diagram for Copy Move Forgery Detection

8. Lexicographically sort the formed matrix.
9. Now check if adjacent rows first value is the equal up to $15^{th}$ decimal digit.
10. If the values match, check the number of times a value is repeating. Also, compute the Euclidean distance between the matching blocks.
11. Apply user specified threshold to eliminate false matches.
12. Create a binary image with one's in the duplicated regions as a result of detecting the forgery.

The previous state of art CMFD process using Hu's invariant moments used moment values up to $4^{th}$ order as features of each block, mainly because of the reason that the value of Hu's invariant moments above $4^{th}$ order generally tend to go beyond $10^{-6}$ units reducing its impact over generation of features. Generation of four invariant moments sufficed the purpose of distinguishing the block among others. In this paper, we propose an algorithm where all the computed Hu's moments are summed to produce one feature value that can distinguish the block from other blocks. Summing up to only $4^{th}$ order moments leads to several false matches. Therefore, in order to reduce false matches to the maximum extent, we produce 7 invariant moments and apply log-polar transform to convert the values beyond $10^{-6}$ units to significant floating values. The accuracy of identifying blocks can further be increased by using 'long format' variables which could display and compute the values generated up to 15$^{th}$ decimal number. Here, if the feature value's matches with one another up to 15$^{th}$ decimal we can have a benefit of doubt that they are duplicated regions. False matches among these are further curbed by calculating Euclidian distance among the matched blocks. The idea here is that, if a cluster of blocks are copied from a region and are duplicated in the same image, the distance between corresponding copied and duplicated block must be the same for every matched pair. A user-specified threshold is applied onto the image to eliminate singular false positives and the remaining matched regions are marked as copy-moved.

Certain feature extraction algorithms such as Scale Invariant Feature Transform (SIFT) or Speeded Up Robust Features (SURF) are commonly used for CMFD purposes. These algorithms mark the features on objects present in the image which provides an advantage of having more robustness towards post processing techniques and geometrical transformations over the pasted region. Since, these algorithms concentrate their key features over objects and drastic pixel flow changes in the image, they often tend to confuse between copy move forgery and genuine similar products in the same image. Using Hu's invariant moments and log-polar transformations to calculate one feature value per block can reduce the chance of false representation over two or more genuine similar products. Hu's moments are sensitive towards the slightest changes in the pixel values which helps' us distinguish between similar products since it's practically impossible to have two or more genuine elements in an image with exactly the same corresponding matching pixels due to the influence over environmental factors, illumination factors and many more.

Moments have well known applications in image processing, computer vision, machine learning and other related fields which are normally used to derive invariants with respect to specific transformation classes.

### A. Hu's invariant moments

For a given two-dimensional continuous function f(x, y), the raw moment of order (p + q) is defined as:

$$M_{pq} = \iint_{-\infty}^{\infty} x^p y^q f(x,y) dx dy \quad (1)$$

All orders in the moments exist if f(x, y) is a continuous bounded function. The distinct moment sequence $\{M_{pq}\}$ can be computed through the function f(x, y) and vice versa. The moments described in equation (1) may not be invariant towards major post processing techniques. However, the required invariant moments can be achieved through the central moments. Central moments are defined as follows:

$$\mu_{pq} = \iint_{-\infty}^{\infty} (x-\bar{x})^p (y-\bar{y})^q f(x,y) dx dy \quad (2)$$

In the above described equation, $(\bar{x}, \bar{y})$ represents the centroid of the image bounded by the function $f(x, y)$. The centroid moments $\mu_{eq}$ are very similar to $\{M_{pq}\}$ whose center is shifted to centroid of the image. This feature of it makes the centroid moments invariant towards several post processing techniques such as rotation and translation. Whereas, scale invariance can be obtained through normalization of the central moments which are defined as follows:

$$\eta_{pq} = \frac{\mu_{pq}}{\mu_{00}^\gamma}, \gamma = \frac{p+q+2}{2} \quad (3)$$

Based on the normalized central moments Hu introduced the following seven moment invariants out of which we use four distinguished moments in order to reduce the dimension of feature vector:

$$\emptyset_1 = \eta_{20} + \eta_{02} \quad (4)$$

$$\emptyset_2 = (\eta_{20} - \eta_{02})^2 + 4\eta_{11}^2$$

$$\emptyset_3 = (\eta_{30} - 3\eta_{12})^2 + (3\eta_{21} - \mu_{03})^2$$

$$\emptyset_4 = (\eta_{30} + \eta_{12})^2 + (\eta_{21} + \mu_{03})^2$$

$$\emptyset_5 = (\eta_{30} - 3\eta_{12})(\eta_{30} + \eta_{12})[(\eta_{30} + \eta_{12})^2 - 3(\eta_{21} + \eta_{03})^2] + (3\eta_{21} - \eta_{03})(\eta_{21} + \eta_{03})[3(\eta_{30} + \eta_{12})^2 - (\eta_{21} + \eta_{03})^2]$$

$$\emptyset_6 = (\eta_{20} - \eta_{02})[(\eta_{30} + \eta_{12})^2 - (\eta_{21} + \mu_{03})^2]$$

$$\emptyset_7 = (3\eta_{21} - \eta_{03})(\eta_{30} + \eta_{12})[(\eta_{30} + \eta_{12})^2 - 3(\eta_{21} + \eta_{03})^2] - (\eta_{30} - 3\eta_{12})(\eta_{21} + \eta_{03})[3(\eta_{30} + \eta_{12})^2 - (\eta_{21} + \eta_{03})^2$$

The above described Hu's invariant moments are generally robust towards noise, JPEG compression, flipping, rotation and rescale geometric transformations.

### IV. EXPERIMENTAL RESULTS

A set of 105 images from the official database MICC- F220, as well as other grey scale images obtained by manual forging process were chosen for the experimental analysis. The images were chosen from diversified environments, varied illumination factors, weather conditions and a varied range over pixel clarity. We chose the standard size of the image to be 256 x 256 and used a constant block size of 8 x 8 in order to attain the best possible results in terms of precision and accuracy. All the experiments were conducted in MATLAB

2016a software with long format variables associated in the program.

*A. Invariance of Hu's moments:*

Among myriad feature extraction algorithms, Hu's invariant moments were selected to serve the purpose due to their effective robustness against several post processing techniques such as JPEG compression, Gaussian noise, rotation at any degree, scaling at any factor and Gaussian blurring. In order to experimentally prove its effectiveness we chose a random 8 x 8 matrix from sample image 1(15:22, 15:22). Its effectiveness can be seen in Table-1 where it is observed that Hu's moments do not drastically change over subjecting the block to several intermediate and post processing techniques. Therefore, this feature of Hu's invariant moments allows us to confidently use them over extracting features of various blocks and match the duplicated regions based on the extracted features.

*B. Feature Vector Dimension:*

Ever since the research on copy move forgery detection started, there is a continuous effort by the researchers to reduce the feature vector dimension in order to reduce space complexity as well as time complexity of the program. In the proposed algorithm, we add all the orders in Hu's invariant moments after applying log polar transforms, to get one feature that represents the particular block of the image. Applying Log-polar transforms would normalize both high and low values to an optimal range which in turn makes these moments highly sensitive to even the minute changes in its pixels. A change of ±0.5 value one pixel could be seen as a significant difference in the calculated sum value as shown in Table-1. Therefore, it can be stated as an effective method to reduce the feature vector dimension to one feature per block. A comparison between the feature vector dimensions of the existing state of art techniques with the proposed algorithm is shown in Table -2.

From Table- 2 it can be inferred that for a 256 x 256 image, if recursive functions are used to calculate the sum of Hu's invariant moments, we can save (62001 * 63) memory locations as compared to DCT analysis. (62001 * 3) memory variables during the execution of the program as compared to the existing technique of CMFD using Hu's invariant moments. Also, the complexity of the program is reduced during lexicographical sorting due to the reduced number of features in the matrix. The Fig3 shows us a visual demonstration of the effectiveness of Algorithm 1.

Table-2. Feature vector dimension.

| Methods | Extraction domain | Block Amount | Feature Dimension |
|---|---|---|---|
| Popescu and Farid [3] | PCA | 62001 | 32 |
| Fridrich et al [1] | DCT | 62001 | 64 |
| Guangjie et al [10] | Hu's invariant moments | 62001 | 4 |
| Proposed algorithm | Sum ( Hu's moments + log polar transforms) | 62001 | 1 |

*C. Performance Evaluation:*

The qualitative results obtained are visually demonstrated in Fig 3. Now we make an attempt to quantitatively measure the effectiveness of the proposed Algorithm 1. To measure the performance Accuracy (A), we define Accuracy A [8] as

$$A = \frac{\text{Number of correctly detected copy – moved pixels}}{\text{Number of pixels actually copy – moved}} \times 100\%$$

The performance accuracy (A) was calculated for different forgery sizes ranging from 10% to 40% of the image, meaning that, 40 % of the image was copied and duplicated in another region of the same image.

Table -3 shows a comparative analysis of the results obtained by the proposed algorithm with the existing state of art techniques with a varied duplication size from 10% to 40%. The highest Performance Accuracy (A) observed and the average Performance Accuracy Values calculated are lucidly displayed for a comprehensive analysis of the effectiveness of proposed algorithm as compared to the existing state of art techniques.

Table-1: Moments after processing through Log-polar transforms are shown below

| Moment order | Original Block | Rotate 153 degrees | Gaussian Blurring | Noise Addition | JPEG Compression | Pixel values with ±0.5 in one of the pixels. |
|---|---|---|---|---|---|---|
| $\emptyset_1$ | 2.5721 | 2.5366 | 2.6221 | 2.7015 | 2.6956 | 2.571629114797040 |
| $\emptyset_2$ | 5.6838 | 5.6799 | 5.6651 | 5.6326 | 5.6541 | 5.685578013766143 |
| $\emptyset_3$ | 10.0654 | 10.1003 | 10.0021 | 10.1602 | 10.0053 | 10.127808708847981 |
| $\emptyset_4$ | 9.0613 | 9.0621 | 9.1023 | 9.0578 | 9.0996 | 9.062931339100622 |
| $\emptyset_5$ | -18.7799 | -18.7659 | -18.7753 | -18.7685 | -18.7712 | -18.819482542125176 |
| $\emptyset_6$ | 12.0346 | 12.0350 | 12.0366 | 12.0451 | 12.0420 | 12.048590406437206 |
| $\emptyset_7$ | 18.6339 | 18.7554 | 18.7124 | 18.6563 | 18.6971 | 18.662871889769104 |

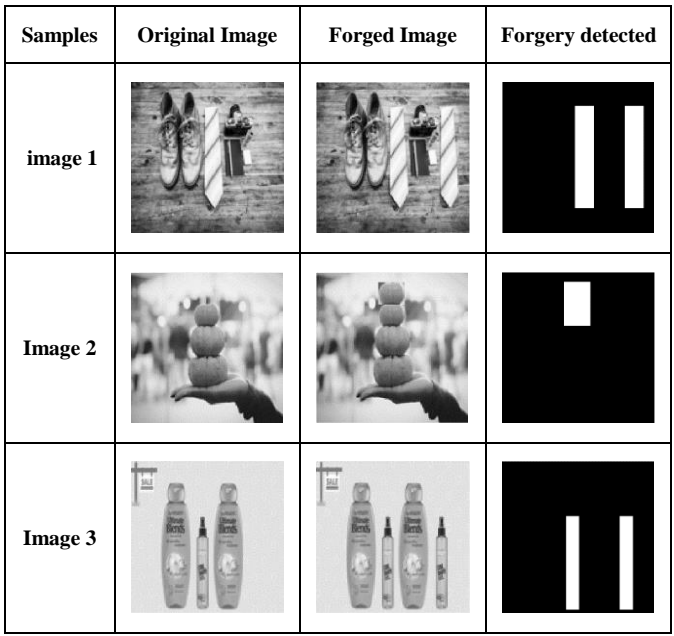

Fig3. Results obtained through Algorithm 1.

Table-3. Performance Accuracy (A) comparison

| Method | Forgery Size (%) | Performance Accuracy (A) – Highest | Performance Accuracy (A) - Average |
|---|---|---|---|
| PCA [3] | 10 | 96.7870 | 96.7588 |
|  | 20 | 96.9130 | 96.9095 |
|  | 30 | 97.1671 | 97.1436 |
|  | 40 | 97.7945 | 97.7645 |
| SVD [4] | 10 | 97.6309 | 97.6092 |
|  | 20 | 98.1880 | 98.1576 |
|  | 30 | 98.4924 | 98.4754 |
|  | 40 | 98.8730 | 98.8311 |
| DCT [1] | 10 | 97.8672 | 97.2254 |
|  | 20 | 97.4396 | 97.4123 |
|  | 30 | 97.6434 | 97.5978 |
|  | 40 | 98.0624 | 98.0232 |
| DWT [5] | 10 | 98.0857 | 98.0838 |
|  | 20 | 98.1490 | 98.1464 |
|  | 30 | 98.2210 | 98.2171 |
|  | 40 | 98.2840 | 98.2780 |
| DyWT [7] | 10 | 98.0027 | 97.9892 |
|  | 20 | 98.3641 | 98.3471 |
|  | 30 | 98.5950 | 98.5455 |
|  | 40 | 98.7889 | 98.7091 |
| Zernike [6] | 10 | 98.8179 | 98.8015 |
|  | 20 | 98.9674 | 98.9372 |
|  | 30 | 99.4017 | 99.3908 |
|  | 40 | 99.4398 | 99.4199 |
| SWT-SVD [8] | 10 | 99.0626 | 99.0362 |
|  | 20 | 99.1391 | 99.1316 |
|  | 30 | 99.4366 | 99.4204 |
|  | 40 | 99.4492 | 99.4307 |
| Proposed Algorithm | 10 | 99.5369 | 99.4549 |
|  | 20 | 99.4569 | 99.4100 |
|  | 30 | 99.4200 | 99.3876 |
|  | 40 | 99.4173 | 99.3821 |

## V. CONCLUSION

Copy-move-forgery is one of the most common forms of passive image forgery. The need to develop advanced, less complex and robust algorithms to curb this kind of forgery is ever increasing. In this paper, we have proposed a novel algorithm to detect copy move forgery with significantly less feature vector dimension. The proposed algorithm is sensitive to even the minute changes of pixel values among the divided blocks and thus capable of lucidly distinguishing genuinely similar objects between copy move forged images. The quantitative and qualitative results obtained delineate the effectiveness of the proposed algorithm. However, the algorithm is less effective when the image is subjected to severe post-processing techniques such as average blurring, rescale and contrast variance etc. In the future, the work can be focused towards developing a single algorithm that is robust towards combinations of multiple post processing techniques.